\documentclass[10pt,twocolumn,letterpaper]{article}

\usepackage{paperstyle}

\usepackage{microtype}

\renewcommand{\paragraph}[1]{\vspace{.5em}\noindent\textbf{#1}}

\usepackage{booktabs}
\usepackage{multirow}
\usepackage{amssymb}
\usepackage{xcolor}
\usepackage{amsthm}
\newtheorem{proposition}{Proposition}

\definecolor{linkblue}{rgb}{0.21,0.49,0.74}
\usepackage{colortbl}
\definecolor{heatblue}{HTML}{0F4C81}%
\usepackage[pagebackref,breaklinks,colorlinks,allcolors=linkblue]{hyperref}

\newcommand{\kAddPoneAILAbutFlat}{4}
\newcommand{\kAddPoneAILAbutGroup}{5}

\newcommand{\kAddPtwoAILAbutFlat}{12}
\newcommand{\kAddPtwoAILAbutGroup}{17}

\newcommand{\kAucBestAll}{0.72}

\newcommand{\kAuthClipShare}{82}

\newcommand{\kCiFailPoneHiA}{87}

\newcommand{\kCiFailPoneLoA}{76}

\newcommand{\kCorrRhoIouOrc}{0.73}
\newcommand{\kCorrRhoIouPone}{0.66}

\newcommand{\kCorrRhoVecOrc}{0.75}
\newcommand{\kCorrRhoVecPone}{0.57}

\newcommand{\kDFourRandN}{20}

\newcommand{\kDesDoneLayerPeeler}{71}

\newcommand{\kDesDoneSuperSVG}{13}

\newcommand{\kDesN}{28}

\newcommand{\kDesNDesignersWord}{three}
\newcommand{\kDesNExcl}{4}

\newcommand{\kDesPoneTrueMethBoth}{four}

\newcommand{\kDesRhoEffCoincMeth}{0.71}

\newcommand{\kDesRhoEffUnownMeth}{0.74}

\newcommand{\kDesRhoEffWandMeth}{\ensuremath{-}0.62}

\newcommand{\kDesTwoItems}{32}

\newcommand{\kDesTwoPoneFalseDone}{24}
\newcommand{\kDesTwoPoneFalseEffort}{3.6}

\newcommand{\kDesTwoPoneFalseN}{45}

\newcommand{\kFailPoneOfOrcA}{82}
\newcommand{\kFailPoneOfOrcC}{42}

\newcommand{\kFailPoneTaskWA}{81}
\newcommand{\kFailPoneTaskWC}{45}

\newcommand{\kFsvFailAll}{41}

\newcommand{\kFsvFailSev}{57}

\newcommand{\kFsvFiles}{20}

\newcommand{\kFsvSevTasks}{9}

\newcommand{\kILAbutDocs}{211}

\newcommand{\kILAbutSameShapes}{210}

\newcommand{\kLinesRefN}{66}

\newcommand{\kNCfgMain}{14}

\newcommand{\kNRhoHiA}{9}

\newcommand{\kNTasksA}{362}
\newcommand{\kNTasksC}{273}

\newcommand{\kPairCap}{500}

\newcommand{\kPertNMaxA}{55}
\newcommand{\kPertNMinA}{50}

\newcommand{\kPertOrcDeletePicosvgA}{99}

\newcommand{\kPertOrcMovePicosvgA}{96}

\newcommand{\kPertPoneFailOrcSplitA}{25}

\newcommand{\kPertPoneFailSplitA}{27}

\newcommand{\kPertPoneGivenPicosvgA}{29}

\newcommand{\kPertPoneGivenShapeSplitA}{56}

\newcommand{\kPoneAIdeogram}{6}

\newcommand{\kPoneAVTracer}{4}

\newcommand{\kPoneCIdeogram}{25}

\newcommand{\kPoneCVTracer}{26}

\newcommand{\kPsnrBaseIdeogram}{34.2}

\newcommand{\kRhoCutDb}{20}

\newcommand{\kRhoDesSame}{0.89}

\newcommand{\kRhoPsnrPoneHiA}{\ensuremath{-}0.20}
\newcommand{\kRhoPsnrPoneHiC}{\ensuremath{-}0.12}

\newcommand{\kSameImgVTFour}{63}
\newcommand{\kSameImgVTOne}{26}

\newcommand{\kSegSuperSVG}{31}

\newcommand{\kSevHi}{88}
\newcommand{\kSevLo}{64}
\newcommand{\kSevOrcTasksDFour}{16}
\newcommand{\kSevOrcTasksDOne}{13}
\newcommand{\kSevShareA}{87}
\newcommand{\kSevShareC}{77}
\newcommand{\kSevShareDFour}{88}
\newcommand{\kSevShareDOne}{64}

\newcommand{\kSinHi}{40}
\newcommand{\kSinLo}{6}
\newcommand{\kSinShareA}{40}
\newcommand{\kSinShareC}{35}

\newcommand{\kSumPoneNotOrcA}{13}
\newcommand{\kSumPoneNotOrcC}{12}

\newcommand{\kTaskRefs}{93}
\newcommand{\kTeaserOrcPass}{two}
\newcommand{\kTeaserPiecesHi}{81}
\newcommand{\kTeaserPiecesLo}{14}

\newcommand{\kUnownSuperSVG}{0.55}

\newcommand{\kVTTuneMaxPone}{7}

\newcommand{\kValDocs}{37}

\newcommand{\kVerAgreeLen}{82}
\newcommand{\kVerAgreeStr}{73}

\newcommand{\kVerN}{76}

\newcommand{\kWandRefDiff}{0.04}

\newcommand{\kWandRefRho}{0.98}

\newcommand{\nCfgWord}{nineteen}

\newcommand{\nMethodsWord}{twelve}

\newcommand{\nTasksAll}{635}

\newcommand{\nValidDocs}{3{,}811}



\title{Same Pixels, Different Documents:\\Measuring Part-Level Editability in Image Vectorization}

\author{Boheng Zhu\textsuperscript{1} \qquad Zehua Wang\textsuperscript{2} \qquad Wei Chen\textsuperscript{1,*} \qquad Ming-Hao Lin\textsuperscript{3}\\
{\normalsize \textsuperscript{1}China University of Mining and Technology \qquad \textsuperscript{2}The University of British Columbia \qquad \textsuperscript{3}Concealed Studio}\\
{\footnotesize \texttt{08250117@cumt.edu.cn}, \texttt{zwang@ece.ubc.ca}, \texttt{chenw@cumt.edu.cn}, \texttt{minghaolin@concealed.studio} \quad \textsuperscript{*}Corresponding author}}

\begin{document}
\twocolumn[{%
\renewcommand\twocolumn[1][]{#1}%
\maketitle
\begin{center}
\vspace{-1.5em}
\includegraphics[width=\linewidth]{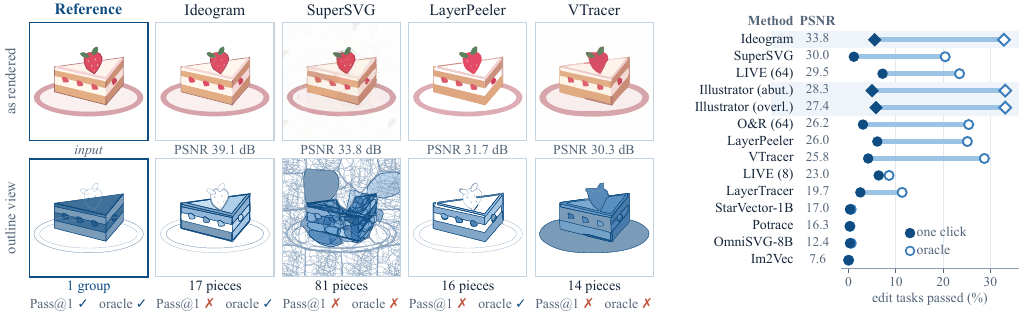}
\captionof{figure}{Same pixels, different documents. Left: an authored
reference and four outputs, rendered and in an editor's outline view, where the
shapes to select for moving the front of the slice are filled, one tint per
selectable object. The reference holds the front as one group. The outputs split
it into \kTeaserPiecesLo{} to \kTeaserPiecesHi{} pieces, so none passes with one
click (Pass@1), whereas the oracle mapping, which collects the pieces itself,
passes \kTeaserOrcPass. Right: authored-set tasks passed with one click and with
the oracle, methods ordered by mean PSNR in dB.}
\label{fig:teaser}
\vspace{1em}
\end{center}
}]
\begin{abstract}
Image vectorization is often evaluated by rendering fidelity, which
cannot distinguish documents that render identically but support different edits.
Edit-based evaluation tests outcomes, but an oracle that gathers a part's
fragments can conceal the cost of selecting them. We introduce Pass@$k$,
which tests whether a prescribed edit can succeed using objects selectable
within $k$ clicks under a specified interaction model. Eight document
diagnostics characterize properties relevant to editing beyond these tasks.
We assess structural
sensitivity through controlled perturbations, practical relevance through
designer operations, and reference dependence through alternative drawings.
We evaluate \nMethodsWord{} open and commercial methods on 211 references,
with \nTasksAll{} edit tasks on \kTaskRefs{} of them. In each of four reference sets, including two drawn by designers,
\kSevLo--\kSevHi\% of oracle-passed tasks on parts composed
of multiple shapes fail at Pass@1, against \kSinLo--\kSinHi\% for single shapes,
pooling counts over configurations.
Higher fidelity does not guarantee higher Pass@1, while selected
diagnostics are associated with designers' effort in our study.
\end{abstract}

\section{Introduction}
\label{sec:intro}

Raster images encode visual content as pixels, whereas vector graphics use
geometric elements with explicit shapes and styles. These elements give designers
direct control over geometry and appearance for precise editing and reuse.
However, many visual assets, including icons, clip art, and generated graphics,
are supplied as raster images. Image vectorization converts them into vector
documents. For design workflows, the quality of a vectorization method depends
on both rendering fidelity and how well the
output structure supports selecting and editing meaningful parts.

Current evaluations focus primarily on rendering fidelity, using metrics such
as PSNR, SSIM, LPIPS, and CLIP
similarity~\cite{wang2004ssim,zhang2018lpips,radford2021clip}. These metrics
cannot distinguish documents that render identically but differ in structure:
grouping elements, for example, can simplify selection without changing their
appearance (\cref{fig:teaser}). Steady improvements in
fidelity~\cite{ma2022live,hirschorn2024optimizereduce,liu2024superpixelsvg,
rodriguez2025starvector,yang2025omnisvg,wu2025layerpeeler,wu2026amodalvec,chen2025easyedit}
therefore do not establish easier editing, even though many methods identify
editability as a goal.

To assess editability more directly, edit-based evaluations compare the
renderings of a reference document and a vectorized output after a prescribed
edit~\cite{park2026compositional}. Editing the output first
requires selecting elements corresponding to the target part.
Park~\etal~\cite{park2026compositional} automatically collect a subset of
output elements to approximate that part. This mapping acts as an oracle,
gathering ungrouped fragments on the user's behalf. It can therefore credit a
successful edit without counting the selections a designer would have to make.

We therefore combine edit outcomes with a selection budget. A task specifies a
reference part and an edit. The output passes at $k$ (\emph{Pass@$k$}) if there
exists a set of elements or groups selectable within $k$ clicks whose edit
satisfies the task's success criteria. This tests feasibility under a specified
selection model. Eight document diagnostics, each tied to
an editor operation, complement this measure by characterizing structures that
can hinder editing.

We evaluate \nMethodsWord{} open and commercial methods on 211 references,
with \nTasksAll{} edit tasks on \kTaskRefs{} of them. To test sensitivity to
structural defects, we apply controlled perturbations that largely preserve
appearance. Moreover, a designer operation study assesses whether the scores
reflect practical editing difficulty. Because tasks depend on how references
define parts, we use references drawn by two designers to examine sensitivity
to alternative decompositions and grouping choices.

Our results reveal a gap between oracle success and Pass@1. For parts
composed of multiple shapes, \kSevLo--\kSevHi\% of oracle-passed tasks
fail at Pass@1 in each of four reference sets (authored, clip, and two drawn
by designers), against \kSinLo--\kSinHi\% for single shapes, pooling counts over
configurations. Higher fidelity does not necessarily imply better
editing performance: Ideogram renders more accurately than VTracer
but shows no clear Pass@1 advantage. In perturbation tests, splitting parts barely affects
the oracle but breaks many one-click edits of single shapes. In the operation
study, selected diagnostics were associated with designers' effort, while higher
PSNR did not indicate easier editing.

\noindent Our contributions are:
\begin{itemize}
\item Pass@$k$, a measure testing edit success under a budget of selection
  clicks, complemented by eight diagnostics.
\item Validation through structural perturbations, designer operations, and
  alternative reference decompositions.
\item A benchmark and comparative assessment of open and commercial
  vectorization methods.
\end{itemize}
References, tasks, method outputs, per-case results, and scoring code are in the supplement and will be released.

\section{Related Work}
\label{sec:related}

\paragraph{Image vectorization.}
Image vectorization methods differ in how they construct document elements.
Classical tracers segment images and fit curves to region
boundaries~\cite{dominici2020polyfit,kopf2011depixelizing,vtracer2023,
selinger2003potrace}, sometimes inferring region order and occluded
boundaries~\cite{chen2024depthvec}. Optimization-based methods fit primitives
through differentiable rasterization~\cite{li2020diffvg}, adding, reducing, or
refining paths~\cite{hirschorn2024optimizereduce,chen2025beziersplatting,
ma2022live,wang2025livss,zhou2024sglive}. Feed-forward networks predict paths
directly~\cite{xu2024deepicon,liu2024superpixelsvg,reddy2021im2vec}; language
models generate SVG code~\cite{rodriguez2025starvector,internsvg2026,
yang2025omnisvg}, sometimes with rendering-based
feedback~\cite{liu2026refinesvg,rodriguez2025rlrf}. Decompose-then-vectorize
pipelines separate or complete raster layers before
tracing~\cite{wu2026amodalvec,park2026compositional,layertracer2025,
wu2025layerpeeler,xu2026semlayer}, addressing the division into parts and, in
some cases, their hidden geometry. These approaches make different structural
choices, raising the question of how to evaluate them for editing.

\paragraph{Evaluating vector graphics.}
Rendering similarity and element counts describe fidelity and document
size~\cite{hirschorn2024optimizereduce,liu2024superpixelsvg,ma2022live,
rodriguez2025starvector}. Structural comparisons include reference geometry in
DeepIcon~\cite{xu2024deepicon}, the share of primitives within a semantic mask
in LIVSS~\cite{wang2025livss}, and agreement between completed parts and
reference shapes in SemLayer~\cite{xu2026semlayer}. These measure correspondence but not edit success or selection cost.

Edit-based evaluations address outcomes more directly. SVG understanding,
generation, and editing benchmarks score rendered outputs or
code~\cite{svgenius2026,cipriano2026svgscore,instruction_svg_edit2025,
svgeditbench2024,svgeditbenchv2_2025,vectorgym2026,internsvg2026,vgbench2024}.
Recent work constructs reference outcomes through controlled
edits~\cite{gupta2026vectorbench,wang2026veriedit,yun2026redesign}, evaluates
edited regions separately from surrounding artwork~\cite{wang2026veriedit},
or replays edits on documents recovered from images~\cite{yun2026redesign}.
For vectorization, Park~\etal~\cite{park2026compositional} evaluate grouping
against annotations and test deletion of annotated front objects. Without
output groups, their oracle identifies each part by the subset of paths that
best reconstructs it, without counting the operations needed to select that
subset. Pass@$k$ additionally requires that a successful edit be achievable
within a selection budget.

\paragraph{Editing effort.}
Operation cost offers a complementary perspective. RGBA layer
decomposition~\cite{chen2025accordion,lungustan2026lade,nie2025ldgd,
wang2026reveallayer,wang2026diffdecompose,yin2026qwenimagelayered,he2026liwi} is typically
evaluated through pixel error and layer overlap. To accommodate multiple valid
decompositions, LayerD~\cite{suzuki2025layerd_true} aligns predicted and
reference layers in paint order, allowing adjacent layers to merge, and reports
quality against merge count as a measure of granularity. ReGroup learns hierarchies of vector primitives that ease
selection~\cite{chaturvedi2021regroup}. Pass@$k$ tests edit success on
vector documents under a selection budget.

\section{Measuring Part-Level Editability}
\label{sec:measures}

We measure whether prescribed edits succeed within a budget of selection clicks
and use document diagnostics to characterize structural features relevant to editing.

\subsection{Why rendering fidelity is insufficient}
\label{sec:structure}

A vector document specifies geometry, appearance, paint order, grouping,
clipping, and fill rules. We distinguish two ambiguities that rendering cannot
resolve: geometry hidden by occlusion and structural representations producing
identical pixels.

Consider elements $e_1,\ldots,e_n$ in paint order under source-over compositing.
Let $\theta_i$, $c_i$, and $a_i(p)\in[0,1]$ denote geometry, color, and effective
coverage, including opacity, at pixel $p$. With other elements and appearance
parameters fixed, the rendered color $C=\mathcal{R}(D)$ is
\begin{equation}
\begin{aligned}
C(p)&=C_{>i}(p)+T_i(p)\bigl[a_i(p)c_i\\
&\qquad +(1-a_i(p))C_{<i}(p)\bigr],\\
T_i(p)&=\prod_{j>i}(1-a_j(p)),
\end{aligned}
\label{eq:composite}
\end{equation}
where $C_{<i}$ composites the background and preceding elements, and $C_{>i}$
is the contribution of subsequent elements.

\begin{proposition}[Occluded geometry]
\label{prop:occlusion}
If $a_j(p)=1$ for some $j>i$, then $\partial C(p)/\partial\theta_i=0$
wherever the derivative exists. Such pixels contribute no gradient with respect
to $\theta_i$ to a differentiable pixelwise reconstruction loss.
\end{proposition}

\begin{proof}
Differentiating the compositing equation gives
\[
\frac{\partial C(p)}{\partial\theta_i}
=T_i(p)\bigl(c_i-C_{<i}(p)\bigr)
\frac{\partial a_i(p)}{\partial\theta_i}.
\]
Complete occlusion gives $T_i(p)=0$.\qedhere
\end{proof}

Coverage changes confined to fully occluded pixels also leave the rendering
unchanged, although moving or deleting the occluder may reveal different
geometry. Priors or constraints shared with visible geometry may determine the
hidden shape; the occluded pixels themselves provide no reconstruction signal
to distinguish alternatives. Partial transparency does not generally give zero gradients.

Independently of occlusion, grouping can preserve appearance while changing
selectable objects. Clipping and explicitly trimmed geometry can render
identically but differ when the clip is released; holes and background-colored
patches can likewise appear identical in a given scene. Whenever
$\mathcal{R}(D_1)=\mathcal{R}(D_2)$, any rendering-only loss or metric with a
fixed target $I$ satisfies $F(\mathcal{R}(D_1),I)=F(\mathcal{R}(D_2),I)$.
This includes pixel and perceptual comparisons, CLIP-based objectives, and
rendering-based feedback for SVG generation~\cite{rodriguez2025rlrf}. Changing
the rendering objective alone cannot distinguish such documents, motivating
tests of edit outcomes, selection costs, and document structure.

\subsection{Defining part-level edit tasks}
\label{sec:edit-test}

A task specifies a reference document $D^\ast$, a target shape or group, and an
operation $e$: \emph{move} translates the part; \emph{delete} removes it;
\emph{recolor} changes its fills while preserving strokes; and \emph{release}
removes its clip. Release tasks target clipped elements. The reference defines
the intended part, but outputs need not reproduce its decomposition or grouping.

Let $A=\mathcal{R}(e(D^\ast))$ be the target rendering and
$B=\mathcal{R}(D^\ast)$ the original. The affected region is
\[
R=\{p:\|A(p)-B(p)\|_\infty>8\},
\]
with RGB values in $[0,255]$. The pair $(A,R)$ specifies the expected outcome
and where the edit should change the image.

\subsection{Pass@\texorpdfstring{$k$}{k}: editing under a selection budget}
\label{sec:passk}

\begin{figure}[t]
\centering
\includegraphics[width=\linewidth]{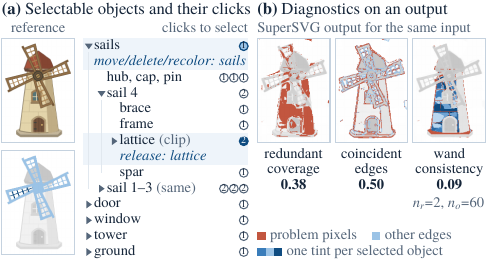}
\caption{(a) Selectable shapes and groups of the authored windmill, with their
click costs. Highlighted targets are the sails group (move, delete, recolor)
and one sail's clipped lattice (release). (b) On SuperSVG's output, coral marks
unowned pixels and boundaries traced by multiple outlines. The Magic Wand
selects $n_r$ reference objects and $n_o$ output objects of the main color.}
\label{fig:method}
\end{figure}

\paragraph{Selection model.}
Vectorization maps $B$ to a document $\hat D$ with rendering
$b=\mathcal{R}(\hat D)$. Under our canvas-based interaction model, selectable
objects are drawable elements or groups outside SVG resources.
Direct element selection costs one click. A group also costs one click if it
has no ancestor partitioning group: a group containing at least two but not all
drawable elements. Groups containing the entire document are treated as layers.
Otherwise, selection costs one initial click on a descendant element plus one
click per group level to the target, counted from its nearest descendant
element. Every hierarchy level offers candidates (\cref{fig:method}a).

\paragraph{Edit success.}
Applying the prescribed edit $e$ to selected objects yields rendering $a$,
without splitting or redrawing geometry. Following VeriEdit~\cite{wang2026veriedit}, we judge the edited
region separately from surrounding artwork. Success requires three conditions:
\emph{fidelity}, mean absolute RGB error between $a$ and $A$ within $R$ at
most 24; \emph{preservation}, mean absolute RGB change between $a$ and $b$
outside a two-pixel dilation of $R$ at most 2; and \emph{localization}, changed
output pixels matching $R$ with precision at least 0.85 and recall at least
0.80 within a two-pixel tolerance. Thresholds were specified by us before scoring any output. On \kVerN{}
sampled oracle-edited outcomes, agreement with one designer was
\kVerAgreeLen\% under lenient judgments and \kVerAgreeStr\% under strict
judgments. Stricter thresholds lower the pass rates but do not change the
comparisons of \cref{sec:results}. Judgment criteria and threshold sensitivity
are in the supplement.

\paragraph{Pass@\texorpdfstring{$k$}{k}.}
A task passes at $k$ if there exists a set of selectable objects, none
containing another, that can be selected within $k$ clicks and whose prescribed
edit $e$ satisfies all three conditions. Release admits only objects carrying their own clip. Counting
clicks includes the cost of reaching nested groups; controlled regrouping shows
how this limits gains from hierarchies unrelated to meaningful parts
(supplement). We report $k=1$ and $k=2$; all reference targets are selectable
within two clicks, so references pass at $k=2$. This measures feasibility under the specified model and assumes that the designer
knows which objects to click, so it is optimistic. Sensitivity to reference part choices is assessed in
\cref{sec:designer-refs}.

\paragraph{Oracle comparison.}
Our implementation of the oracle mapping of Park~\etal~\cite{park2026compositional} greedily collects
output elements matching the reference part's visible region, with the
thresholds and tolerance of the localization condition. It lifts them to the
outermost group whose visible elements are all selected. Selection has no click
budget, but edits face the same success criteria. Greedy matching can miss
selections found by Pass@$k$, so the oracle is not an upper bound on Pass@$k$:
summed over methods, \kSumPoneNotOrcA{} authored-set and \kSumPoneNotOrcC{}
clip-set tasks pass Pass@1 but not the oracle. The
fraction of oracle-passed tasks failing at Pass@1 characterizes the selection
gap, not additional clicks or completion time.

\paragraph{Search and approximation.}
We prune candidates using bounds on where edits can change pixels and render
at most \kPairCap{} pairs per task. Both pruning and the pair limit can omit
successful selections, so computed Pass@$k$ scores may underestimate the true
pass rates under the existential definition. On
\kValDocs{} documents, pruned and exhaustive searches agreed on all task
verdicts despite pruning some successful candidates. This is empirical agreement,
not a completeness guarantee. Search details, validation results, and tasks
affected by the pair limit are in the supplement.

\subsection{Structural diagnostics}
\label{sec:diagnostics}

Pass@$k$ tests task feasibility; the following structural inventory and eight
diagnostics characterize document properties beyond those tasks. Individually, the diagnostics predict one-click failures only weakly (AUC at most
\kAucBestAll; supplement).

\paragraph{Structural inventory.}
\label{sec:inventory}
For each reference and output, we count drawable elements and record five descriptors: content clips, partitioning groups (\cref{sec:passk}),
holes, translucent elements, and stacking overlap. A
\emph{content clip} removes at least 1\% of the clipped element's coverage,
measured on the element alone with and without the clip. Holes are unfilled regions
enclosed by a filled path under its fill rule. For filled regions
$E_1,\ldots,E_n$ in paint order, stacking overlap is
\begin{equation}
O=\frac{\sum_i\left|E_i\cap\bigcup_{j>i}E_j\right|}
        {\sum_i|E_i|}.
\label{eq:overlap}
\end{equation}
It measures the fraction of painted area covered by later elements; $O=0$ for
a disjoint tiling. Multiple structures can be valid, so these descriptors have
no universally preferred values. References specify narrower choices of parts,
clipped regions, and holes. Thresholds are in the supplement.

\paragraph{Stacking and boundaries.}
\emph{Redundant coverage} is the share of painted pixels whose color remains
unchanged whichever single element is hidden (the unowned pixels in
\cref{fig:method}b); moving visible elements can expose this coverage.
\emph{Coincident edges} measures the share of visible color boundaries traced
by multiple element outlines, potentially requiring coordinated edits. Both use
per-element coverage and visibility information.

\paragraph{Path and color complexity.}
\emph{Micro-segments} measures the share of curve segments shorter than
1.5 pixels with a tangent change greater than $45^\circ$ at a join: small, sharp
features that can complicate path adjustment. The \emph{segment ratio} and
\emph{palette ratio} divide output segment and distinct fill-color counts by
reference counts, describing relative complexity without prescribing optimal
counts.

\paragraph{Color selection and line representation.}
Two diagnostics follow interviews with four designers, only one of whom joined
the operation study (\cref{sec:designers}). \emph{Wand consistency} emulates
Illustrator's fill-color selection at tolerance 32 on the 0--255 scale. For each
reference color group, selected object counts $n_r$ and $n_o$ in the reference
and output give $\min(n_r,n_o)/\max(n_r,n_o)$, multiplied by the IoU of selected
pixel regions; scores are averaged by reference-region area. \emph{Line
recovery} measures the proportion of reference stroke-only pixels also
represented by stroke-only output elements, distinguishing centerlines from
filled outlines that require different reshaping operations.

\paragraph{Line-art fidelity.}
\emph{Black-and-white IoU} compares input and output dark-pixel sets after
grayscale thresholding below 128, complementing line recovery with rendered
line-art agreement. Detailed computations and threshold sensitivity are in the
supplement.

\section{Benchmark and Evaluation Protocol}
\label{sec:benchmark}

We evaluate \nMethodsWord{} methods on 211 reference documents, with
\nTasksAll{} edit tasks on \kTaskRefs{} of them.

\subsection{Reference documents}
\label{sec:data}

\paragraph{Base set.}
The 173 base documents comprise 118 FreeSVG icons and clip-art files (CC0),
excluding gradients, patterns, filters, masks, embedded images, and text; and
55 documents that we authored, with written specifications of intended parts, clips,
holes, and groups. Named shapes or groups specify their edit targets.
Public-file selection criteria are in the supplement.

\paragraph{Clip set.}
Content clips (\cref{sec:inventory}) are uncommon in the surveyed public
collections. This rarity describes what the collections contain, not how often
designers use clipping, so a separate set of 38 documents with native clipping
tests them: 32 Noto Emoji (Apache-2.0) and six Flat UI icons (MIT). Documents were
de-duplicated against the base set and chosen for varied clipping uses;
collection statistics and selection details are in the supplement.

\subsection{Task construction}
\label{sec:task-construction}

Benchmark tasks use only the 55 authored and 38 clip-set documents, independently of
vectorization outputs. Move, delete, and recolor target shapes or groups at
depth one or two, restricted to named parts in authored documents. Their visible
region---pixels changing when the part is hidden---must cover 100 pixels to half the canvas. Move translates the part by 20 pixels in the first direction
that keeps it on the canvas; delete requires at least 50 pixels of other content
beneath the part; recolor
rotates fill hues by $120^\circ$, preserving strokes. Release targets clipped
elements at any depth. Tasks require an affected region $R$
(\cref{sec:edit-test}) of at least 64 pixels. Fixed rules select at most two
tasks per family and reference (supplement).

\subsection{Compared methods and protocol}
\label{sec:methods}

The ten open methods span five paradigms (\cref{sec:related}):
classical tracers (Potrace~\cite{selinger2003potrace}, VTracer~\cite{vtracer2023}),
optimization (LIVE~\cite{ma2022live}, O\&R~\cite{hirschorn2024optimizereduce}),
feed-forward networks (Im2Vec~\cite{reddy2021im2vec},
SuperSVG~\cite{liu2024superpixelsvg}), language models
(StarVector-1B~\cite{rodriguez2025starvector}, OmniSVG-1.1-8B~\cite{yang2025omnisvg}),
and decompose-then-vectorize pipelines (LayerPeeler~\cite{wu2025layerpeeler},
LayerTracer~\cite{layertracer2025}). Other such pipelines~\cite{wu2026amodalvec,park2026compositional,xu2026semlayer} lacked
public implementations in October 2026. We also evaluate two commercial tools
whose methods are unpublished: Adobe Illustrator's Image Trace and Ideogram
Vector.

The \kNCfgMain{} main configurations include LIVE with 8 and 64 paths and O\&R with 64. Illustrator's
Image Trace runs with auto-grouping in both of its modes: abutting, which cuts
shapes apart, and overlapping, which stacks them. Open methods retain repository configurations without
benchmark tuning. Versions, checkpoints, preprocessing, presets, and additional
path budgets are documented in the supplement.

All methods receive the same $256\times256$ RGB reference rendering on white
and retain their own preprocessing, including resizing. A fixed Chromium build renders all documents. Valid outputs must parse, render, and be
nonblank. Means use valid outputs; invalid outputs fail all edit tasks.

\section{Validating the Measures}
\label{sec:validation}

Before using the measures to compare vectorization methods, we examine
what their scores capture. Controlled perturbations test whether
Pass@$k$ and the diagnostics respond to targeted structural changes
while largely preserving appearance (\cref{sec:perturb}). We then
examine how these scores relate to designers' effort in practical
editing tasks (\cref{sec:designers}). Because Pass@$k$ evaluates parts
defined by a reference, we also use designer-drawn references to test
how alternative part decompositions and grouping choices affect pass
rates and method rankings (\cref{sec:designer-refs}).

\begin{table}[t]
\centering
\caption{Controlled perturbations of the 55 authored references. PSNR:
median against the unperturbed rendering; perturbed documents below
30\,dB, unchanged documents and failed transformations are excluded,
leaving \kPertNMinA--\kPertNMaxA{} documents per row. Orac., P@1: tasks passed (\%); P@1 is not conditioned on passing on
the reference. Sel.: select-same-color IoU (supplement); other
columns as in \cref{tab:diagnostics} (medians). Shading grows with
the change from the reference row.}
\label{tab:perturb}
\footnotesize
\setlength{\tabcolsep}{2.3pt}
\begin{tabular}{@{}l rrrrrrrr@{}}
\toprule
Perturbation & PSNR & Orac. & P@1 & Unown. & Coinc. & Micro & Seg.$\times$ & Sel. \\
\midrule
Reference & -- & 96 & 84 & 0.009 & 0.12 & 0.000 & 1.0 & 1.00 \\
\midrule
flatten & 46.7 & \cellcolor{heatblue!18}82 & \cellcolor{heatblue!45}25 & 0.007 & \cellcolor{heatblue!15}0.22 & \cellcolor{heatblue!11}0.044 & \cellcolor{heatblue!30}3.4 & \cellcolor{heatblue!13}0.93 \\
cut out & 34.2 & \cellcolor{heatblue!45}40 & \cellcolor{heatblue!45}9 & 0.000 & \cellcolor{heatblue!45}0.81 & \cellcolor{heatblue!21}0.172 & \cellcolor{heatblue!42}6.7 & \cellcolor{heatblue!14}0.92 \\
split & 50.1 & 95 & \cellcolor{heatblue!14}75 & \cellcolor{heatblue!14}0.096 & \cellcolor{heatblue!11}0.17 & \cellcolor{heatblue!10}0.029 & \cellcolor{heatblue!19}1.8 & 1.00 \\
duplicate & 34.4 & \cellcolor{heatblue!45}24 & \cellcolor{heatblue!20}68 & \cellcolor{heatblue!45}0.876 & \cellcolor{heatblue!45}1.00 & 0.000 & \cellcolor{heatblue!20}2.0 & 1.00 \\
color jitter & 59.1 & 96 & 84 & 0.009 & 0.12 & 0.000 & 1.0 & 1.00 \\
subdivide & 43.5 & 94 & 84 & 0.009 & 0.12 & 0.009 & \cellcolor{heatblue!41}6.4 & 1.00 \\
extra anchors & 99.0 & 96 & 84 & 0.009 & 0.12 & \cellcolor{heatblue!25}0.226 & \cellcolor{heatblue!13}1.3 & 1.00 \\
\bottomrule
\end{tabular}
\end{table}

\subsection{Controlled perturbations}
\label{sec:perturb}

We apply seven perturbations to both edit sets to test responses to
structural changes (\cref{tab:perturb}).
\emph{Flatten} removes groups and converts clips into geometry;
\emph{split} divides each opaque, fill-only shape into two
same-colored pieces while preserving groups. \emph{Cut out} additionally
removes geometry hidden by opaque shapes after flattening;
\emph{duplicate} overlays each opaque shape with an identical copy.
\emph{Color jitter} perturbs solid fills by at most one level per
RGB channel; \emph{subdivide} splits each segment into four; and
\emph{extra anchors} inserts duplicate anchors at sharp corners.

Expected responses were specified before scoring. Some transformations
altered anti-aliased boundaries. After inspecting PSNR but before
examining the other measures, we retained only perturbed documents
with PSNR of at least 30\,dB against the original rendering.
Unchanged documents and failed transformations were excluded;
full settings, clip-set results and results without the PSNR rule are in the supplement.

Splitting exposes a gap between oracle success and one-click
editability. On the authored set, \kPertPoneFailSplitA{} tasks pass
Pass@1 on the reference but fail after splitting; the oracle still
passes \kPertPoneFailOrcSplitA{} of them. Among tasks passing Pass@1
on the reference, \kPertPoneGivenShapeSplitA\% of tasks targeting
single shapes remain successful after splitting, compared with
nearly all tasks targeting groups, whose pieces remain selectable
together.

Flattening retains Pass@1 for only \kPertPoneGivenPicosvgA\% of tasks
that pass Pass@1 on the reference, although the oracle passes
\kPertOrcMovePicosvgA\% of move and
\kPertOrcDeletePicosvgA\% of delete tasks. Flattening also removes
clips, so its overall effect cannot be attributed solely to grouping.

Duplication and cutting out can also prevent a correct edit, even
with oracle selection (\cref{tab:perturb}). Moving or deleting only
one duplicate leaves another in place; cutting out removes hidden
geometry that an edit may expose.

The diagnostics respond to changes beyond the tested edit outcomes.
As expected, redundant coverage increases
with duplication, coincident edges with cutting out and flattening,
micro-segments with extra anchors, and the segment ratio with
subdivision. For jitter, subdivision and extra anchors, every task
that initially passes Pass@1 still passes after perturbation in both sets.

Some perturbations affect several diagnostics: duplication also
increases coincident edges, and flattening and cutting out increase
micro-segments. Color jitter leaves median select-same-color IoU
unchanged, contrary to the prediction, but increases the palette
ratio, a response not specified in advance (supplement).

\subsection{Designer operation study}
\label{sec:designers}

To relate the scores to practical editing effort,
\kDesNDesignersWord{} graphic designers edited reference documents and
outputs of VTracer, LayerPeeler and SuperSVG in Illustrator. Each item
specified one edit on one file: moving, deleting, recoloring, Select
Same, reshaping or stroking an outline. In the first round, four source
images yielded \kDesN{} items after excluding \kDesNExcl{} reference
items imported incompletely by Illustrator. Each trial allowed two
minutes, with no penalty for abandoning it; designers recorded
completion, time and effort (1 easy--5 very hard). Tasks were checked
to be easy on the references. Participants gave written informed consent,
were paid and are reported anonymously (ethics details in the supplement).

Higher PSNR did not indicate easier editing in this study. All
reference items were completed, whereas SuperSVG, the most accurate
of the three methods, had only \kDesDoneSuperSVG\% of trials completed,
compared with \kDesDoneLayerPeeler\% for LayerPeeler.
LPIPS, for which lower is better, increased with effort, so the finding about PSNR does not extend
to all image metrics.

Among method outputs, greater redundant coverage and more coincident edges were
associated with greater mean effort, while higher wand consistency
was associated with less effort ($\rho=\kDesRhoEffUnownMeth$,
$\kDesRhoEffCoincMeth$ and $\kDesRhoEffWandMeth$, respectively).
These associations are consistent with the work of manipulating
overlapping elements, coordinating shared boundaries and selecting
colors. Several items share each document, so treating
all items as independent would overstate the evidence (full results
in the supplement).

A second round on four new source images with the same designers
reproduced the ordering of file types by pooled completion and effort.
The color-selection result also held for Designers~2 and~3, who had
not proposed the wand criterion: they abandoned every Select Same
trial on SuperSVG but completed those on the other file types.
This agrees with SuperSVG's low wand consistency and provides
evidence beyond the criterion's proposers (supplement).

Support for Pass@1 is narrower. Only \kDesPoneTrueMethBoth{} method
items pass across the two rounds, and every trial on them
was completed with an effort of 1. In the second round, designers
also completed \kDesTwoPoneFalseDone{} of
\kDesTwoPoneFalseN{} trials on items whose associated tasks failed
Pass@1, with mean effort \kDesTwoPoneFalseEffort. Thus passing
coincided with easy completion in this sample, while failure did not
preclude completion using the editor's broader interactions. Some
study operations use benchmark edits on the same part as proxies:
reshaping, for example, uses a move-task verdict (supplement).
The small number of passing items and the difference in operations
limit claims about general predictive performance.

\subsection{References drawn by designers}
\label{sec:designer-refs}

We test dependence on reference-defined parts by rescoring fixed
method outputs against two designers' redrawings.
Designer~1 from the operation study redrew four public base-set images;
Designer~4, who took no other part in the study, redrew the same four
and \kDFourRandN{} additional sampled images. All redrawings met
the 25\,dB fidelity criterion; sampling rules and prespecified
expectations are in the supplement.

Different decompositions define different edits. On the shared bee
image, Designer~1 drew a continuous body beneath the stripes, whereas
Designer~4 used abutting pieces. Deleting a stripe therefore reveals
different geometry. Across the four shared images, each designer's
drawings fail some of the other's tasks even with oracle mapping
(supplement). The score thus reflects compatibility with the requested edits.

Absolute pass rates change, while the multi-shape selection gap persists.
With identical outputs on
the four shared images, VTracer's Pass@1 is \kSameImgVTOne\% under
Designer~1's references and \kSameImgVTFour\% under Designer~4's
(release tasks excluded). Across the designers' full reference sets,
\kSevShareDOne\% and \kSevShareDFour\% of oracle-passed tasks on
parts composed of multiple shapes (\kSevOrcTasksDOne{} and \kSevOrcTasksDFour{} tasks)
fail at Pass@1, pooling counts over method configurations. On \kFsvFiles{} public files drawn at random and scored
with their authors' parts, \kFsvFailAll\% of oracle-passed tasks fail at Pass@1,
and \kFsvFailSev\% on multi-shape parts (\kFsvSevTasks{} tasks; supplement).

Relative rankings are more stable: on the four shared images,
methods' Pass@1 ranks correlate at $\rho=\kRhoDesSame$ between the
two authors' references. Wand consistency also agrees closely for
the same outputs ($\rho=\kWandRefRho$; mean absolute difference
\kWandRefDiff). Only the ranking expectation held fully; the others
held partly or failed (supplement). The observed stability
remains limited to the tested references.

\section{Results}
\label{sec:results}

We first examine what selection budgets reveal about the outputs, then
whether rendering fidelity reflects this performance, and finally document
properties relevant to editing.

\subsection{Editing under a selection budget}
\label{sec:res-editing}

\begin{table}[t]
\centering
\caption{Edit tasks passed (\%) with at most one or two clicks (P@1, P@2) and
with the oracle mapping, which collects every element of the part (Orac.;
\cref{sec:passk}). PSNR: mean over the task documents. Reference: the pipeline
on the references, where P@2 holds by construction. Invalid outputs fail.}
\label{tab:editing}
\footnotesize
\setlength{\tabcolsep}{2.6pt}
\begin{tabular}{@{}l rrrr|rrrr@{}}
\toprule
& \multicolumn{4}{c|}{Authored (\kNTasksA{} tasks)} & \multicolumn{4}{c}{Clip set (\kNTasksC{} tasks)} \\
Method & PSNR & P@1 & P@2 & Orac. & PSNR & P@1 & P@2 & Orac. \\
\midrule
Reference & -- & 84 & 100 & 96 & -- & 99 & 100 & 98 \\
\midrule
Potrace & 16.3 & 0 & 0 & 0 & 10.7 & 1 & 1 & 1 \\
VTracer & 25.8 & 4 & 11 & 29 & 27.3 & 26 & 33 & 40 \\
LIVE (8) & 23.0 & 6 & 9 & 9 & 24.9 & 10 & 16 & 20 \\
LIVE (64) & 29.5 & 7 & 15 & 23 & 33.2 & 7 & 16 & 24 \\
O\&R (64) & 26.2 & 3 & 8 & 25 & 28.2 & 6 & 13 & 27 \\
Im2Vec & 7.6 & 0 & 0 & 0 & 7.9 & 0 & 0 & 0 \\
SuperSVG & 30.0 & 1 & 2 & 20 & 30.8 & 1 & 1 & 6 \\
StarVector-1B & 17.0 & 0 & 1 & 1 & 15.4 & 6 & 7 & 8 \\
OmniSVG-8B & 12.4 & 0 & 1 & 1 & 8.8 & 1 & 1 & 2 \\
LayerPeeler & 26.0 & 6 & 15 & 25 & 28.7 & 25 & 31 & 40 \\
LayerTracer & 19.7 & 2 & 6 & 11 & 17.7 & 13 & 16 & 17 \\
\midrule
Illustrator (abut.) & 28.3 & 5 & 17 & 33 & 32.0 & 23 & 28 & 31 \\
Illustrator (overl.) & 27.4 & 6 & 20 & 33 & 29.0 & 28 & 36 & 38 \\
Ideogram & 33.8 & 6 & 13 & 33 & 35.4 & 25 & 29 & 36 \\
\bottomrule
\end{tabular}
\end{table}

\paragraph{Oracle success can conceal one-click failures.}
\Cref{tab:editing} compares Pass@1, Pass@2 and oracle success.
Pooling task outcomes over evaluated configurations,
\kFailPoneOfOrcA\% of oracle-passed authored-set tasks fail at
Pass@1 (95\% interval \kCiFailPoneLoA--\kCiFailPoneHiA\%, bootstrap over references), compared
with \kFailPoneOfOrcC\% on the clip set. Weighting each task equally gives
\kFailPoneTaskWA\% and \kFailPoneTaskWC\%, respectively. Thus a successful oracle
edit often does not translate into success under a one-click budget.

The gap is larger for parts composed of several shapes. Among
oracle-passed tasks, \kSevShareA\% and \kSevShareC\% of these tasks
fail at Pass@1 on the authored and clip sets, respectively, compared
with \kSinShareA\% and \kSinShareC\% for single shapes.
Multi-shape targets account for a larger share of non-release tasks
in the authored set than in the clip set (supplement), consistent
with its larger aggregate gap. Oracle and Pass@1 rankings are also less closely aligned on the
authored set than on the clip set (supplement): accounting for
selection changes the comparison between methods.

\paragraph{Grouping changes access to existing shapes.}
\label{sec:res-tools}
Illustrator's abutting mode provides a concrete example. With and
without auto-grouping, \kILAbutSameShapes{} of \kILAbutDocs{}
document pairs contain the same shapes. On the authored set,
auto-grouping raises Pass@2 from \kAddPtwoAILAbutFlat\% to
\kAddPtwoAILAbutGroup\%, but Pass@1 only from
\kAddPoneAILAbutFlat\% to \kAddPoneAILAbutGroup\%; oracle rates
remain similar. Most groups are nested, so reaching them can exceed
one click. This comparison illustrates why the selection model counts
access through the hierarchy, rather than treating every group as
equally accessible.

\subsection{Fidelity and editability}
\label{sec:res-fidelity}

\begin{table}[t]
\centering
\caption{Base-set overview (173 inputs). Valid: nonblank outputs that parse
and render. PSNR: mean over valid outputs. Group, Hole: shares of valid
outputs with partitioning groups or holes. Full results are in the supplement.}
\label{tab:base-overview}
\small
\setlength{\tabcolsep}{4.2pt}
\begin{tabular}{@{}lrrrr@{}}
\toprule
Method & Valid & PSNR$\uparrow$ & Group & Hole \\
\midrule
Potrace & 168 & 14.37 & 0.00 & 0.55 \\
VTracer & 173 & 26.08 & 0.00 & 0.31 \\
LIVE (8) & 173 & 23.28 & 0.00 & 0.00 \\
LIVE (64) & 173 & 29.87 & 0.00 & 0.00 \\
O\&R (64) & 173 & 26.63 & 0.00 & 0.00 \\
Im2Vec & 173 & 7.46 & 0.00 & 0.00 \\
SuperSVG & 173 & 29.08 & 0.00 & 0.00 \\
StarVector-1B & 16 & 15.00 & 0.13 & 0.25 \\
OmniSVG-8B & 166 & 11.07 & 0.00 & 0.28 \\
LayerPeeler & 173 & 25.61 & 0.00 & 0.62 \\
LayerTracer & 173 & 18.28 & 0.00 & 0.60 \\
\midrule
Illustrator (abut.) & 173 & 29.45 & 0.98 & 0.97 \\
Illustrator (overl.) & 173 & 28.25 & 0.99 & 0.00 \\
Ideogram & 173 & 34.19 & 0.00 & 0.10 \\
\bottomrule
\end{tabular}
\end{table}

Higher fidelity does not guarantee better Pass@1.
Ideogram has the highest mean PSNR on the base set
(\kPsnrBaseIdeogram\,dB; \cref{tab:base-overview}).
On the task documents,
it also renders more accurately than VTracer, but their Pass@1 rates
are \kPoneAIdeogram\% versus \kPoneAVTracer\% on the authored set
and \kPoneCIdeogram\% versus \kPoneCVTracer\% on the clip set
(\cref{tab:editing}). The 95\% intervals for the paired differences
include zero on both sets (supplement), so the comparison does not
resolve a clear Pass@1 advantage for Ideogram.

Across all configurations, mean PSNR and Pass@1 are positively
rank-correlated. Poor reconstruction can itself prevent an edit from
meeting the fidelity criterion, providing a possible explanation for
this association. Among the \kNRhoHiA{} configurations above
\kRhoCutDb\,dB, however, the correlations are
$\rho=\kRhoPsnrPoneHiA$ and $\kRhoPsnrPoneHiC$ on the authored and
clip sets. Correlations computed within fixed references also
average near zero (supplement). Additional VTracer experiments
improve fidelity through tuning or larger inputs but raise
authored-set Pass@1 to at most \kVTTuneMaxPone\% (supplement).
Together, these comparisons show why fidelity alone is insufficient
to assess the editing performance of these outputs.

\subsection{Document structure and editing consequences}
\label{sec:res-structure}

\begin{table}[t]
\centering
\caption{Diagnostics on the base set (medians over valid outputs; B/W: mean).
Unown.: redundant coverage; Coinc.: coincident edges; Micro: micro-segment rate;
Seg.$\times$, Pal.$\times$: segments and fill colors relative to the reference;
Wand: wand consistency; Lines: line recovery, over the \kLinesRefN{} references
with lines; B/W: black-and-white IoU (\cref{sec:diagnostics}). Lower is better
for the first three columns and higher for the last three.}
\label{tab:diagnostics}
\footnotesize
\setlength{\tabcolsep}{1.45pt}
\begin{tabular}{@{}l rrrrr|rrr@{}}
\toprule
Method & Unown. & Coinc. & Micro & Seg.$\times$ & Pal.$\times$ & Wand & Lines & B/W \\
\midrule
Reference & 0.01 & 0.15 & 0.00 & 1.0 & 1.0 & 1.00 & 1.00 & 1.00 \\
\midrule
Potrace & 0.00 & 0.00 & 0.00 & 0.5 & 0.1 & 0.00 & 0.00 & 0.95 \\
VTracer & 0.03 & 0.45 & 0.02 & 3.6 & 3.1 & 0.45 & 0.00 & 0.84 \\
LIVE (8) & 0.00 & 0.05 & 0.00 & 0.2 & 0.8 & 0.13 & 0.00 & 0.68 \\
LIVE (64) & 0.04 & 0.25 & 0.00 & 2.0 & 3.2 & 0.20 & 0.00 & 0.86 \\
O\&R (64) & 0.05 & 0.17 & 0.00 & 2.5 & 3.6 & 0.22 & 0.00 & 0.81 \\
Im2Vec & 0.00 & 0.00 & 0.00 & 0.6 & 0.4 & 0.00 & 0.00 & 0.06 \\
SuperSVG & 0.55 & 0.63 & 0.29 & 31.0 & 34.6 & 0.05 & 0.00 & 0.86 \\
StarVector-1B & 0.00 & 0.01 & 0.00 & 0.9 & 0.9 & 0.20 & 0.07 & 0.58 \\
OmniSVG-8B & 0.04 & 0.31 & 0.03 & 2.7 & 0.3 & 0.02 & 0.00 & 0.22 \\
LayerPeeler & 0.00 & 0.27 & 0.18 & 5.6 & 2.7 & 0.23 & 0.00 & 0.82 \\
LayerTracer & 0.06 & 0.57 & 0.13 & 15.3 & 4.9 & 0.09 & 0.00 & 0.43 \\
\midrule
Illustrator (abut.) & 0.00 & 0.99 & 0.07 & 2.4 & 4.0 & 0.51 & 0.00 & 0.88 \\
Illustrator (overl.) & 0.00 & 0.08 & 0.05 & 2.1 & 4.0 & 0.50 & 0.00 & 0.88 \\
Ideogram & 0.00 & 0.44 & 0.03 & 1.6 & 1.1 & 0.54 & 0.00 & 0.91 \\
\bottomrule
\end{tabular}
\end{table}

\paragraph{No output contains a content clip.}
Across \nValidDocs{} valid outputs from \nCfgWord{} configurations,
including supplementary path budgets, no output contains a content
clip, whereas \kAuthClipShare\% of the authored references and every clip-set
reference contain one. Correspondingly, every method fails all release tasks, which
test whether removing a clip reveals the intended artwork.
These failures lower absolute pass rates uniformly within each set,
but do not change method rankings or the conditional selection gap,
because the oracle also passes none of them. Holes and partitioning groups, by contrast, do appear in
some outputs, but never in those of the methods that fit renderings
(\cref{tab:base-overview}; \cref{sec:structure}).

\paragraph{Other document properties can complicate editing.}
The diagnostics (\cref{tab:diagnostics}) characterize properties
beyond the tested edits. In SuperSVG's median output,
redundant coverage is \kUnownSuperSVG{} and the segment ratio is
\kSegSuperSVG, indicating overlapping coverage and complex paths.
Multiple outlines also trace many visible boundaries in accurate
outputs, potentially requiring coordinated changes when reshaping
them. Color selection and line representation introduce further
differences: SuperSVG has much lower wand consistency than VTracer
and Ideogram, and most configurations have zero median line
recovery. These observations identify document properties relevant
to editing; they do not attribute each Pass@1 failure to a specific
diagnostic.

\section{Discussion and Limitations}
\label{sec:discussion}

\paragraph{Implications.}
Pass@$k$ measures support for reference-defined edits, rather than
agreement with a unique document decomposition. The different valid
authorings in \cref{sec:designer-refs} illustrate why absolute scores
depend on editing intent. Comparisons should therefore use shared
references and distinguish part types, with fidelity and diagnostics
providing complementary information.

Method design should also account for access to parts. The grouping
comparison (\cref{sec:res-tools}) illustrates that useful hierarchies
matter beyond group presence. Retaining a clip preserves a native
operation that baking it into geometry removes. When such alternatives
render identically, fidelity alone cannot distinguish them
(\cref{sec:structure}), motivating investigation of structural priors
and objectives that evaluate edits.

\paragraph{Limitations.}
The benchmark contains 211 references, with edits tested on
\kTaskRefs{} of them. Its 55 authored documents emphasize clips,
holes and groups: these examples probe structural capabilities,
rather than estimate how often deficits occur in production artwork.
Fixed edit magnitudes and the small study of alternative references
leave other editing intents and document complexities untested.

We evaluate only $k=1,2$ in a click model that assumes the user knows
which output objects to select. Scores use fixed success thresholds and omit
discovery time, alternative selection tools and repair; color selection
is assessed separately by wand consistency. Release specifically
requires native clipping. Pruning and the pair cap can miss successful
selections, so computed scores may underestimate feasibility under
the model; the greedy oracle is not an upper bound on Pass@$k$.

The operation study comprises \kDesN{} and \kDesTwoItems{} items in
two rounds with the same \kDesNDesignersWord{} designers. Items share
documents, some operations use proxy verdicts, and only
\kDesPoneTrueMethBoth{} method items pass Pass@1, limiting evidence
that Pass@1 predicts practical editing difficulty. Method results concern
the tested releases and settings, without measuring variability across
seeds. Several methods lacked public implementations at evaluation
time. Other structure measures were compared only as reimplementations on the
authored set: across configurations, the VeC of LIVSS~\cite{wang2025livss} and a part
IoU after SemLayer~\cite{xu2026semlayer} correlate more with the oracle
($\rho=\kCorrRhoVecOrc$ and $\kCorrRhoIouOrc$) than with Pass@1 ($\kCorrRhoVecPone$ and
$\kCorrRhoIouPone$; supplement).

\paragraph{Conclusion.}
Rendering similarity alone cannot establish part-level editability.
Evaluating prescribed edits under a selection budget exposes gaps
that oracle matching can conceal: in all four reference sets, most
oracle-passed tasks on parts composed of multiple shapes fail at Pass@1 under our protocol. Higher fidelity does not guarantee higher
Pass@1, and no evaluated output contains a content clip.

\section*{Acknowledgments}
This research was funded by the National Key-Supported College Student Innovation
Training Program of China University of Mining and Technology (Project
No.~202610290072). We would like to express our sincere gratitude to the students
of the school's Artificial Intelligence Club for their enthusiastic support, to all
the designers who took part in this work for their time and effort, and to Rui Shen
for dedicated assistance in polishing the manuscript.

\section*{LLM Usage Statement}
An LLM-based assistant was used under the author's direction: to check
related work; to debug the code; to check the reported numbers against
the result files; and to improve the text. The authors conceived and
directed the research, made the design decisions, conducted the designer
study, produced the code and writing, revised the text, and take full
responsibility for the content. The methods evaluated that use language models
(StarVector, OmniSVG, and LayerPeeler, which calls Gemini) are part of
the benchmark and are described in \cref{sec:methods}.

{
    \par\ifdefined\nolinenumbers\nolinenumbers\fi
    \small
    \bibliographystyle{ieeenat_fullname}
    \bibliography{refs}
}
\end{document}